\documentclass{article}
\usepackage{iclr2027_conference,times}
\usepackage[T1]{fontenc}
\usepackage[table,dvipsnames]{xcolor}
\usepackage{graphicx}
\usepackage{booktabs}
\usepackage{microtype}
\usepackage[most]{tcolorbox}
\usepackage{hyperref}
\usepackage{url}
\usepackage[noabbrev,nameinlink]{cleveref}
\hypersetup{hidelinks}

\usepackage{amsmath, amssymb, amsthm, mathtools}
\usepackage{enumitem}
\usepackage{xspace}
\usepackage{tabularx}
\usepackage{tikz}
\usetikzlibrary{arrows.meta, positioning, fit, shapes.geometric, calc}

\theoremstyle{definition}

\tcolorboxenvironment{definition}{colback=blue!4, colframe=blue!4, arc=1mm, boxrule=0pt}
\tcolorboxenvironment{designprinciple}{colback=red!4, colframe=red!4, arc=1mm, boxrule=0pt}

\title{Large Knowledge Model: A Knowledge\\ Foundation for Agentic Science at Scale}
\author{
Yuan~Huang\textsuperscript{1,2,*}, Sihan~Hu\textsuperscript{2,1}, Hongyu~Gu\textsuperscript{1} \\
Chao~Ma\textsuperscript{1}, Jiaxing~Zhang\textsuperscript{1}, Zhiyong~Zou\textsuperscript{1} \\
Caiyu~Fan\textsuperscript{1}, Yan~Xiao\textsuperscript{1}, Mingjun~Xu\textsuperscript{1} \\
Chenyu~Xie\textsuperscript{1}, Mingzhen~Ju\textsuperscript{1}, Zhehao~Ma\textsuperscript{1} \\
Qi~Zhang\textsuperscript{1}, Baozong~Wang\textsuperscript{3}, Yu~Li\textsuperscript{2,4} \\
Zhiyuan~Yao\textsuperscript{2,4}, Ruoxue~Liao\textsuperscript{1,*}, Xinyu~Li\textsuperscript{5,*} \\
Linfeng~Zhang\textsuperscript{1,5,*}, Kun~Chen\textsuperscript{2,*}, Weinan~E\textsuperscript{5,6,7,*} \\
\textsuperscript{1}DP Technology, Beijing, China \\
\textsuperscript{2}Institute of Theoretical Physics, Chinese Academy of Sciences, Beijing, China \\
\textsuperscript{3}International Center for Quantum Materials, School of Physics, Peking University, Beijing, China \\
\textsuperscript{4}Lanzhou Center for Theoretical Physics, Lanzhou University, Lanzhou, China \\
\textsuperscript{5}AI for Science Institute, Beijing, China \\
\textsuperscript{6}School of Mathematical Sciences, Peking University, Beijing, China \\
\textsuperscript{7}Center for Machine Learning Research, Peking University, Beijing, China \\
\textsuperscript{*}Corresponding authors \\
\texttt{huangyuan@dp.tech}; \texttt{chenkun@itp.ac.cn} \\
\texttt{liaorx@dp.tech}; \texttt{lixy@dp.tech} \\
\texttt{zhanglf@dp.tech}; \texttt{weinan@math.pku.edu.cn}
}

\iclrfinalcopy

\begin{document}
\maketitle

\begin{abstract}
Agentic science envisions many autonomous agents investigating concurrently
while building on a shared, evolving body of scientific knowledge. This requires
a knowledge foundation that supports high-concurrency access, preserves
traceable and reusable reasoning, and grows incrementally. We propose the
\textbf{Large Knowledge Model (LKM)}, a growing, agent-native knowledge
foundation that provides a general representation of scientific knowledge
across disciplines. LKM organizes the scientific literature into
reasoning graphs, with claims as the core nodes and associated reasoning chains
that make explicit how premises and evidence support conclusions. These
source-grounded objects are persistent and addressable; cross-paper links
organize them into aligned question, workflow, and evidence views. Newly extracted
papers extend the foundation incrementally while preserving existing object
identities. Building on this foundation, we develop an \textbf{agent-native,
reasoning-aware scientific retrieval system} that retrieves claims together with
their reasoning chains and sources, enabling agents to inspect and reuse the
evidence underlying scientific conclusions. Across benchmarks, agents using LKM
retrieve more evidence, cite more faithfully, and answer scientific questions
more accurately: LKM nearly
doubles the known supporting and contradicting evidence retrieved on SciFact-Open
(818 versus 443 claim--paper pairs), reasoning graphs raise citation F1 on ScholarQABench
by more than 5 points over the same retrieved papers, and LKM retrieval improves a fixed answering model
by 9.3, 4.2, and 14.7 points over no retrieval on ChemBench, PubMedQA, and SciBench. LKM lays the
foundation for a scientific ecosystem in which AI scientists not only recall
accumulated knowledge but also extend it, returning new questions, workflows, and
evidence to a memory that every subsequent investigation can build on.
\end{abstract}
\begin{center}\small\url{https://lkm.bohrium.com/web/en}\end{center}

\section{Introduction}
\label{sec:introduction}

Agentic science promises a research paradigm in which many autonomous agents pursue investigations concurrently and build on accumulated scientific knowledge. Agents can already search the literature and synthesize evidence~\citep{he2025pasa,skarlinski2024language,asai2026synthesizing}; sustained investigation further requires them to establish what previous work has shown, how its conclusions were reached, and when they apply. As investigations multiply, agents need a shared, evolving scientific record that they can inspect and reuse across tasks and disciplines.

The existing record was written for human readers. A paper's argument, from its premises and evidence to its conclusions and the conditions under which they hold, is spread across prose, equations, figures, and references, and document or passage retrieval leaves agents to reconstruct it for every query~\citep{lewis2020retrieval,asai2026synthesizing}. The central challenge is a knowledge representation that spans disciplines with diverse terminology, methods, and forms of evidence, supports agent access and reasoning, and remains faithful to the source literature. Scientific arguments offer a common basis: claims can be expressed as propositions connected to the premises, evidence, and reasoning that support them under stated conditions. Reasoning graphs built from these relations make scientific knowledge accessible to agents while retaining the context and provenance needed to interpret it.

\begin{figure*}[t]
  \centering
  \includegraphics[width=\linewidth]{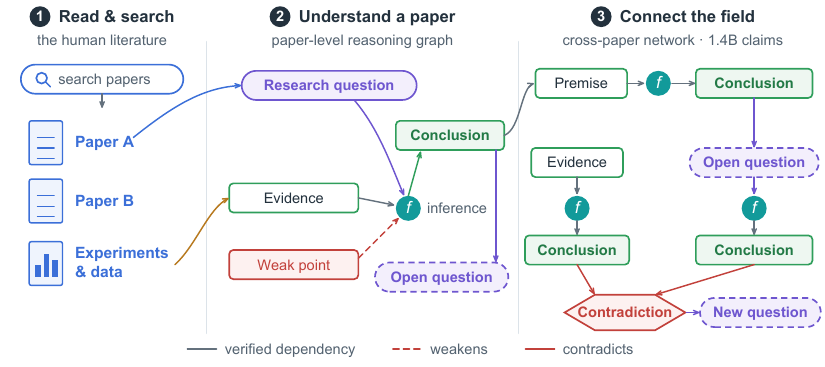}
  \caption{\textbf{From reading papers to new questions.} An agent searches and reads the literature (left). LKM extracts each paper into a reasoning graph in which the research question, evidence, and weak points enter an inference that reaches a conclusion and leaves open questions (center). Across papers, conclusions become premises of other inferences, open questions are taken up by later work, and contradictions between conclusions raise new questions (right).}
  \label{fig:global-reasoning-landscape}
\end{figure*}

We present the \emph{Large Knowledge Model} (LKM), a growing, agent-native knowledge foundation for science, built through four operations (\Cref{fig:memory-overview}). \emph{Extraction} turns each paper into a source-grounded reasoning graph whose questions, claims, and reasoning chains are addressable objects linked by typed dependencies. \emph{Organization} joins recurring objects across papers and arranges them into a \emph{Scientific Reasoning Landscape} with three aligned views: a Question Landscape of research problems and open directions, a Workflow Landscape of reusable research procedures, and an Evidence Landscape relating conclusions to their support, disagreement, and conditions. \emph{Retrieval} lets an agent find a question, claim, or procedure and follow it to its premises, reasoning, and source. \emph{Updates} are incremental: each newly extracted paper extends LKM without rewriting existing object identities. Because every agent reaches the same persistent objects, what one investigation retrieves, another can inspect, compare, and extend. \Cref{fig:global-reasoning-landscape} traces this path from reading and searching papers to the contradictions and open questions that surface across a field.

LKM is deployed today. Its reasoning graphs span the major scientific disciplines and hold about 1.4 billion vectorized claims; it organizes nearly one million cross-paper question families, and answers requests from 50 concurrent clients with a median latency of 740~ms. Agents access it through an open API, a command-line interface, and agent skills, and researchers through web interfaces. Agents that build on LKM do better scientific work. LKM nearly doubles the known supporting and contradicting evidence retrieved for scientific claims on SciFact-Open (818 versus 443 claim--paper pairs within a 50-paper budget). On ScholarQABench, giving the writer the reasoning graphs of the same retrieved papers raises citation F1 by 5.3 and 5.1 points. With the answering model fixed, LKM retrieval improves accuracy over no retrieval by 9.3, 4.2, and 14.7 points on ChemBench, PubMedQA, and SciBench.

This foundation points toward a new mode of scientific progress. Open questions in the Question Landscape can seed new investigations; workflow families supply tested procedures to adapt; and the Evidence Landscape is designed to expose which observation would resolve a disagreement. The next step closes the loop: agents will return their own observations, failures, and conclusions to LKM as new reasoning objects, so that they accumulate knowledge collectively rather than rediscovering it in isolation.

Our contributions are:
\begin{enumerate}[leftmargin=*, itemsep=2pt, topsep=2pt]
\item \textbf{An agent-native representation of scientific knowledge} that turns papers into source-grounded reasoning graphs with addressable questions, claims, and multi-step reasoning (\Cref{sec:reasoning-graphs}).
\item \textbf{A knowledge foundation at the scale of the literature}, organized into a Scientific Reasoning Landscape of questions, workflows, and evidence, and updated incrementally to about 1.4 billion claims (\Cref{sec:global-landscape}).
\item \textbf{A reasoning-aware retrieval system and evidence that agents benefit from it}: agents using LKM retrieve more evidence, cite more faithfully, and answer scientific questions more accurately (\Cref{sec:evaluation}).
\end{enumerate}

\section{Related Work}
\label{sec:related-work}

\paragraph{Scientific retrieval and synthesis.}
Dense retrieval and retrieval-augmented generation connect language models to large corpora~\citep{karpukhin2020dense,lewis2020retrieval}. Scientific systems build on this foundation: OpenScholar combines scientific retrieval with citation-backed synthesis~\citep{asai2026synthesizing}, PaperQA2 uses language agents to search, synthesize, and resolve conflicting findings~\citep{skarlinski2024language}, and PaSa and PaSaMaster perform iterative agentic paper search~\citep{he2025pasa,du2026towards}. These systems retrieve documents or passages at query time. LKM supplies them with a persistent knowledge foundation in which the reasoning of each paper is already explicit, so retrieval returns scientific objects that can be expanded, compared, and cited.

\paragraph{Structured representations of science.}
Scholarly corpora and knowledge graphs organize bibliographic links, entities, and contributions~\citep{lo2020s2orc,jaradeh2019orkg,luan2018multi,jain2020scirex}. Argument-centered representations capture how knowledge is established: argumentative zoning labels rhetorical roles~\citep{teufel2002summarizing}, nanopublications and micropublications attach assertions to provenance, evidence, and challenges~\citep{groth2010nanopublication,clark2014micropublications}, and \emph{Linking Scholarly Contents} builds intra- and inter-paper argumentation graphs~\citep{song2022linking}. \emph{The Discovery Engine} envisions navigable knowledge landscapes for agent-oriented discovery~\citep{baulin2025discovery}. LKM realizes argument-level structure across the literature of the major scientific disciplines, organizes it around research questions and multi-step reasoning, and aligns it into question, workflow, and evidence landscapes that agents can query.

\paragraph{Graph-augmented retrieval.}
GraphRAG extracts entity--relation graphs and summarizes their communities for query-focused synthesis~\citep{edge2024local}; LightRAG combines entity graphs with keyword and vector retrieval~\citep{guo2025lightrag}. Their graphs are organized around entities, relation descriptions, and topical communities. LKM makes scientific roles first-class instead: questions, premises, reasoning steps, conclusions, and open directions, connected by directed inferential relations such as \texttt{premise\_of} and \texttt{concludes}. This representation exposes how a finding is established, under which conditions it holds, and how later work reuses or challenges it.

\section{Extracting Reasoning Graphs from Papers}
\label{sec:reasoning-graphs}

The unit of LKM is not a document or a passage but a scientific reasoning object. We represent each paper $p$ as a typed graph with inference factors, $\mathcal{G}_p=(\mathcal{V}_p,\mathcal{F}_p,\mathcal{E}_p)$, where $\mathcal{V}_p$ contains the paper's scientific objects, $\mathcal{F}_p$ contains inference factors that map an ordered set of premises to a conclusion, and $\mathcal{E}_p$ contains typed relations among objects and factors. An agent can enter this graph through a question, a claim, or a procedure and follow the dependencies relevant to its task.

\subsection{Reasoning Objects}
\label{sec:reasoning-primitives}

\Cref{tab:reasoning-objects} lists the object types. A \emph{question} states what an investigation seeks to resolve; a \emph{claim} is a proposition that serves as a premise or conclusion; a \emph{reasoning chain} is an ordered sequence of steps $\sigma=(s_1,\ldots,s_T)$ explaining how a factor's premises yield its conclusion. \emph{Settings} record the experimental or domain conditions under which claims hold, and \emph{highlights} and \emph{weak points} record where an argument is strongest and where it most needs scrutiny. Relations include \texttt{addresses} (question to claim), \texttt{premise\_of} (premise to factor), \texttt{concludes} (factor to conclusion), \texttt{subproblem\_of}, \texttt{highlight\_of}, and \texttt{weakpoint\_of}. Every object keeps its source references, so a claim retrieved from LKM always leads back to the paper that established it.

\begin{table}[t]
\centering
\caption{\textbf{Reasoning objects in a paper-level graph.}}
\label{tab:reasoning-objects}
\small
\begin{tabularx}{\linewidth}{@{}lX@{}}
\toprule
\textbf{Object} & \textbf{Role in LKM} \\
\midrule
Question & Research problem, subproblem, or open question that frames the paper's claims \\
Claim & Decontextualized proposition that appears as a premise or conclusion \\
Inference factor & Maps an ordered premise set to a conclusion and records the method used \\
Reasoning chain & Ordered steps attached to a factor, explaining how the conclusion is reached \\
Setting & Experimental or domain conditions under which claims and chains apply \\
Highlight / weak point & Strongest supporting element and load-bearing vulnerability of an argument \\
\bottomrule
\end{tabularx}
\end{table}

\subsection{Construction}

LKM builds each graph in stages from the paper's normalized full text. It first extracts the paper-level problem, its conclusions and their subproblems, and the open questions the paper leaves unresolved. For each conclusion, it reconstructs the reasoning chain and identifies the premises and evidence that carry the argument, together with load-bearing weak points and supporting highlights. It then rewrites conclusions and premises as self-contained propositions that remain meaningful outside their original paragraph, preserving their identifiers and source references. After structural validation, these objects form the paper's reasoning graph. Extraction is performed by DeepSeek-V4-Flash.

We audited extraction fidelity on 100 randomly sampled papers comprising 3,904 extracted objects. An LLM judge compared each object with the full text, and domain experts reviewed every flagged case; the confirmed hallucination rate is 0.97\%. Most confirmed errors are semantic: a correct number attached to the wrong object, a miscount, a reversed direction, or an unsupported attribution to an experiment.

\begin{figure}[t]
    \centering
    \includegraphics[width=\linewidth]{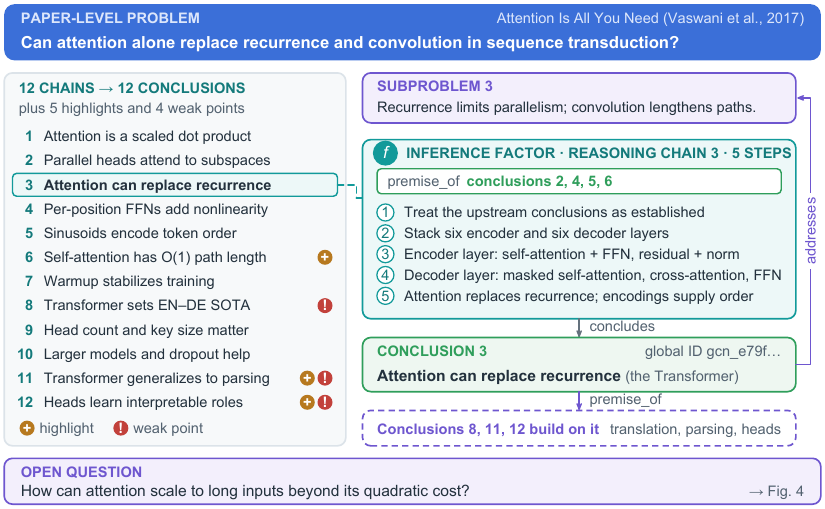}
    \caption{\textbf{Anatomy of an extracted paper.} The reasoning graph LKM stores for \emph{Attention Is All You Need}~\citep{vaswani2017attention}. Left: its twelve chains, each named by its conclusion, with highlights (+) and weak points (!). Right: chain~3 expanded; conclusions 2, 4, 5, and 6 are its premises, and five steps conclude the Transformer, which addresses its subproblem and supports conclusions 8, 11, and 12.}
    \label{fig:paper-level-reasoning-graph}
\end{figure}

\Cref{fig:paper-level-reasoning-graph} shows the graph extracted from \emph{Attention Is All You Need}. The paper yields twelve reasoning chains, each reaching one conclusion that addresses its own subproblem, together with five highlights and four weak points. Scaled dot-product attention is the premise of multi-head attention, and multi-head attention, the feed-forward network, positional encoding, and the complexity analysis enter the inference that concludes the Transformer architecture, which in turn is a premise of the translation, parsing, and attention-head results. The inference expands into five ordered steps; a weak point marks that the translation results come from single runs without confidence intervals, and a highlight marks that the architecture transfers to parsing. The quadratic cost of self-attention is stored as an open question, which is where the cross-paper traversal of \Cref{sec:questions-workflows-evidence} begins.

\subsection{Identity and Retrieval}

LKM keeps each paper's own objects while giving recurring objects a shared identity across the literature. A public claim's identity is determined by its object type, its decontextualized proposition text $x_v$, and its parameters in canonical order, and is materialized as a SHA-256 hash, the global ID shown in \Cref{fig:paper-level-reasoning-graph}. Objects with the same identity attach to one global representative that retains all of its source-paper occurrences; factors are integrated by matching their premises, conclusion, and type. Semantic similarity never merges identities. It forms a separate layer of neighborhoods and clusters used for discovery and for the cross-paper views of \Cref{sec:global-landscape}.

For retrieval, each public claim is embedded as $\mathbf{z}_v=\operatorname{Embed}(x_v)\in\mathbb{R}^{512}$ with Qwen3-Embedding-8B~\citep{zhang2025qwen3embedding}, using its Matryoshka support to request 512-dimensional vectors directly; reasoning chains are embedded from their ordered step text. Claim vectors let agents find \emph{what} was established, and chain vectors let them find \emph{how} it was established. Hybrid retrieval combines these vectors with BM25 over titles and full text. Because retrieved identifiers resolve to graph objects, a single request can move from a semantic match to the claim's premises, reasoning steps, settings, and source paper.

\section{Organizing Knowledge at Scale}
\label{sec:global-landscape}

\begin{figure}[t]
    \centering
    \includegraphics[width=\linewidth]{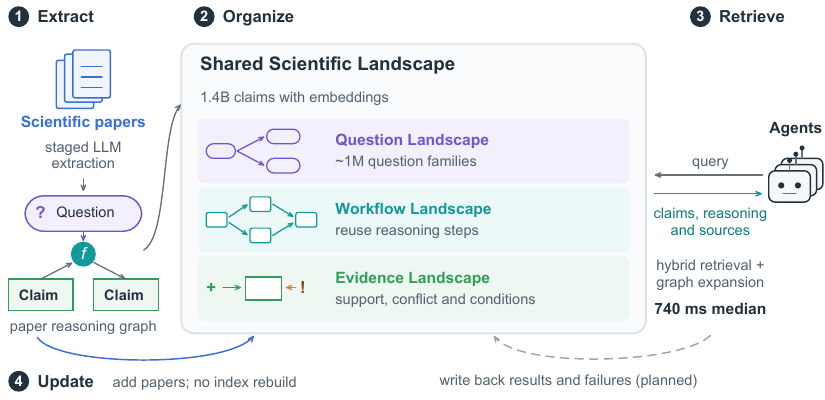}
    \caption{\textbf{LKM as a knowledge foundation for agentic science.} Papers are (1) extracted into reasoning graphs and (2) organized into Question, Workflow, and Evidence Landscapes. Agents (3) retrieve claims together with their reasoning and sources, and the foundation is (4) updated incrementally with each new paper. The dashed path marks the planned write-back of results produced by agents.}
    \label{fig:memory-overview}
\end{figure}

Paper-level graphs expose the reasoning within individual investigations. Organization connects that reasoning across the literature, so that an agent can see what a field has asked, how it has investigated those questions, and what the resulting evidence establishes. \Cref{fig:memory-overview} summarizes the resulting foundation: papers are extracted, organized into three aligned landscapes, retrieved by agents, and continually updated as new papers arrive.

\subsection{Scale and Incremental Updates}
\label{sec:lkm-scale-search}

The deployed system contains reasoning graphs spanning all of arXiv, bioRxiv and ChemRxiv preprints, and journals in mathematics, physics, chemistry, astronomy, earth and life sciences, engineering, information science, and environmental science. They hold about 1.4 billion vectorized public claims, complemented by abstract representations for a broader set of papers. Local and global records, canonical bindings, embeddings, and clusters are persisted in a cloud-native analytical database built on ClickHouse, and the vector index is served by 16 machines with 32 CPU cores and 256~GB of memory each. Under 50 concurrent clients sustained for 10 minutes, the production search endpoint answers with a median end-to-end latency of 740~ms and a 95th-percentile latency of 980~ms, fast enough to sit inside an agent's reasoning loop.

LKM is updated one paper at a time. When a paper has been extracted, each public object is resolved by its exact identity: it joins an existing global representative if one exists and creates a new one otherwise. Existing bindings are never rewritten, and only the objects introduced or matched by the new paper enter the embedding update. The cost of adding a paper is therefore proportional to its own content rather than to the size of the corpus, and a newly published result becomes retrievable without rebuilding the index.

\subsection{The Scientific Reasoning Landscape}
\label{sec:questions-workflows-evidence}

Over the corpus $\mathcal{P}$ of extracted papers, we define the \emph{Scientific Reasoning Landscape} as $\mathcal{L}=(\mathcal{Q},\mathcal{W},\mathcal{B},\mathcal{R})$, where $\mathcal{Q}$ is a set of cross-paper question families, $\mathcal{W}$ a set of workflow families, $\mathcal{B}$ a set of claim-centered evidence analyses, and $\mathcal{R}$ the relations connecting them. All three views are built from the same paper-level objects, so an inquiry can move from a question to the procedures used to study it and on to the evidence behind their conclusions (\Cref{fig:memory-overview}, center).

\paragraph{Question Landscape.}
The Question Landscape organizes the literature by the problems that motivate it. Question families are formed by clustering the research questions that papers address together with the open questions they leave unresolved; an LLM then summarizes each family's shared question, its main variants, and the papers that investigate it, with every summary linked to its member objects. On an arXiv snapshot of 34.4 million questions, this produces 988,894 families that cover 99.5\% of the papers, ranging from focused recurring problems to broad research themes (\Cref{app:question-family-construction}). An open question thus becomes an entry point to everything the literature has tried.

\paragraph{Workflow Landscape.}
The Workflow Landscape organizes recurring ways of producing scientific knowledge~\citep{belhajjame2015suite,kulkarni2018annotated}. A workflow family clusters reasoning chains that follow a common procedure and summarizes its inputs and outputs, reasoning skeleton, task variants, and technique slots, the alternative techniques that can fill each step. Each family also records whether it is a dry, wet, or mixed investigation and how its results are verified. For an agent, a workflow family is reusable planning material: a sequence of scientific purposes, the choices available at each step, and source examples of how others instantiated it.

\paragraph{Evidence Landscape.}
The Evidence Landscape turns claim retrieval into evidence analysis~\citep{wadden2020fact}. Given a question, the Evidence Engine issues complementary queries, retrieves claims, expands their reasoning chains, and organizes the result as a source-linked evidence graph. Evidence is separated into supporting, opposing, conditional, insufficient, and methodologically divergent items, each tied to its premises, settings, and source. Contradiction groups record which claims are in tension and which conditions explain the disagreement, and research-gap records identify the observation that would resolve it.

\begin{figure}[t]
    \centering
    \includegraphics[width=\linewidth]{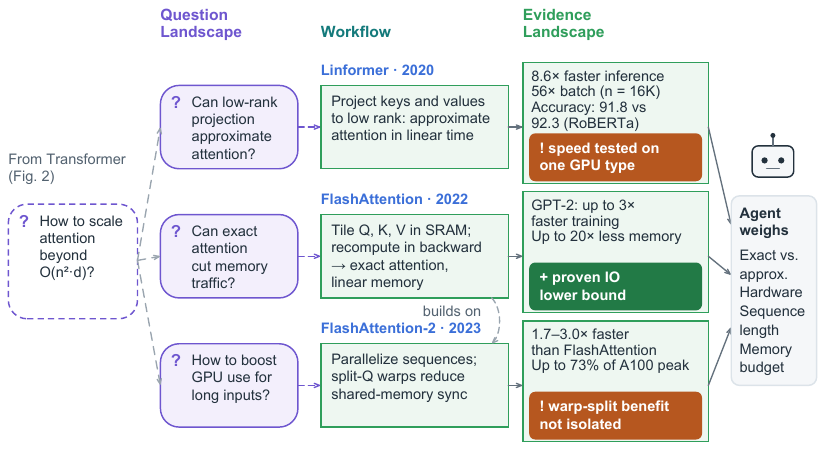}
    \caption{\textbf{An agent traversing the reasoning landscape.} Starting from the open question left by the Transformer paper (\Cref{fig:paper-level-reasoning-graph}), LKM surfaces three routes to efficient attention. Each route connects a subquestion to a method and to its evidence, with highlights (+) and weak points (!) attached, so the agent can choose an approach by exactness, hardware, and sequence length.}
    \label{fig:landscape-attention-case}
\end{figure}

\Cref{fig:landscape-attention-case} shows how an agent traverses the landscape. It starts from the open question stored with the Transformer paper, how attention can scale beyond its quadratic cost, and the Question Landscape divides it into three subquestions answered by different papers~\citep{wang2020linformer,dao2022flashattention,dao2023flashattention2}. Linformer asks whether a low-rank projection can approximate attention; FlashAttention asks whether exact attention can cut memory traffic; FlashAttention-2, which builds on it, asks how to raise GPU utilization for long inputs. Each subquestion leads to a method in the Workflow Landscape and to the evidence behind its claims: the speedups and accuracy each paper reports, a proven IO lower bound for FlashAttention, and weak points such as Linformer's speed being measured on a single GPU type.

\subsection{Interfaces for Agents}

Agents reach LKM through an OpenAPI service, a command-line interface, and agent skills that expose the same operations as tool calls. The core operations are hybrid search over claims, questions, and abstracts; search over complete reasoning chains; retrieval of a paper's full reasoning graph; tracing why a claim holds through its premises and steps; and PDF extraction for papers not yet in LKM. Researchers use the same objects through web interfaces (\Cref{app:interfaces}).

\section{Experiments}
\label{sec:evaluation}

We test whether agents do better scientific work when they recall reasoning rather than documents. The experiments follow three capabilities: discovering the evidence that bears on a scientific claim (\Cref{sec:exp-evidence}), writing answers whose citations support what they state (\Cref{sec:exp-attribution}), and finding papers and answering knowledge-intensive questions (\Cref{sec:exp-qa}). All experiments query the deployed LKM service.

\subsection{Evidence Discovery}
\label{sec:exp-evidence}

Judging a scientific claim requires finding the papers that support or contradict it. We evaluate this with SciFact-Open, an open-domain claim-verification benchmark~\citep{wadden2022scifactopen,wadden2020fact}, using its 279 claims and all 460 human-annotated positive claim--paper pairs. To account for relevant papers outside the benchmark's original annotation pool, we pool candidates from the released retriever and LKM and admit a new positive pair only when a DeepSeek judge (\texttt{deepseek-v4.1-flash}; SUPPORT and CONTRADICT thresholds 0.80 and 0.90) and MultiVerS~\citep{wadden2022multivers} agree; human annotations always take precedence. The resulting reference set contains 1,125 known-positive pairs. Each system retrieves 50 papers per claim.

\begin{table}[t]
\centering
\caption{\textbf{Evidence discovery on SciFact-Open at 50 papers per claim.} monoT5 reranks LKM's candidates without changing the retrieved set.}
\label{tab:scifact-evidence-discovery}
\small
\begin{tabularx}{\linewidth}{@{}Xrrr@{}}
\toprule
\textbf{Metric} & \textbf{Released} & \textbf{LKM} & \textbf{LKM+monoT5} \\
\midrule
Known-positive pairs retrieved & 443 & \textbf{818} & \textbf{818} \\
Evidence recall & 39.38\% & \textbf{72.71\%} & \textbf{72.71\%} \\
\quad SUPPORT recall & 34.35\% & \textbf{75.80\%} & \textbf{75.80\%} \\
\quad CONTRADICT recall & 47.36\% & \textbf{67.82\%} & \textbf{67.82\%} \\
Claim hit rate & \textbf{89.73\%} & 85.27\% & 85.27\% \\
$\mathrm{MRR}@50$ & \textbf{0.7174} & 0.5364 & 0.5707 \\
\bottomrule
\end{tabularx}
\end{table}

Within the same 50-paper budget, LKM retrieves 818 known-positive pairs, compared with 443 for the released retriever (\Cref{tab:scifact-evidence-discovery}). The claim-paired recall gain is 33.3 points (95\% bootstrap interval 24.6--42.1). The gain holds for both directions of evidence: SUPPORT recall rises from 34.4\% to 75.8\%, and CONTRADICT recall from 47.4\% to 67.8\%, so an agent recalling from LKM sees more of the evidence that could overturn a claim, not only the evidence that confirms it. Retrieval breadth and ranking play complementary roles: LKM finds 540 known positives in its first 20 ranks and 278 more in ranks 21--50, and a downstream reranker such as monoT5 moves 81 of those later positives into the top 20, adapting the enlarged evidence pool to an agent's attention budget.

\subsection{Attributable Scientific Synthesis}
\label{sec:exp-attribution}

Long-form scientific answers are only as trustworthy as their citations. We evaluate synthesis and attribution on ScholarQA-CS (100 questions) and ScholarQA-Multi (108 questions) from ScholarQABench~\citep{asai2024openscholar} with an OpenScholar-style pipeline: retrieval, reranking, drafting from the top 10 papers, self-feedback, and citation repair, with GPT-4o as the writer. We compare three LKM configurations: LKM Search, LKM Search with reranking, and LKM Graph with reranking, which shares its candidates, reranking, and 10-paper writing budget with LKM Search + rerank and additionally gives the writer the reasoning graphs of the top three papers. Answer quality follows the benchmark's native protocols, the GPT-4o rubric Overall score on ScholarQA-CS and the Prometheus score for organization, relevance, and coverage on ScholarQA-Multi; citation precision, recall, and F1 are computed with the benchmark's AttrScore evaluator. Published results of OpenScholar, PaperQA2~\citep{skarlinski2024language}, and SciRAG~\citep{ding2025scirag} serve as references.

\begin{figure}[t]
    \centering
    \includegraphics[width=\linewidth]{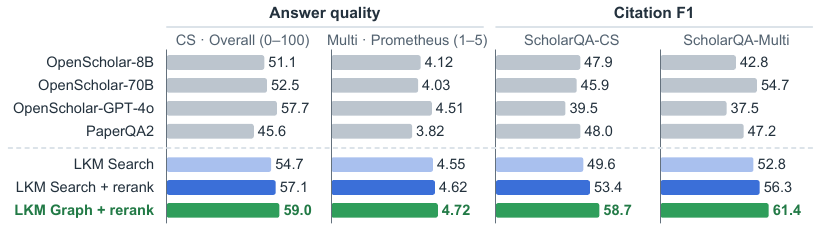}
    \caption{\textbf{Scientific synthesis on ScholarQABench.} Answer quality (left; GPT-4o rubric Overall on ScholarQA-CS, Prometheus on ScholarQA-Multi) and citation F1 (right) for three LKM configurations and published results of OpenScholar and PaperQA2 (gray). LKM Graph + rerank differs from LKM Search + rerank only in giving the writer the reasoning graphs of the top papers.}
    \label{fig:scholarqa}
\end{figure}

LKM Graph obtains the highest answer quality (59.0 on ScholarQA-CS, 4.72 on ScholarQA-Multi) and citation F1 (58.7 and 61.4) among our configurations on both subsets (\Cref{fig:scholarqa}). These scores also exceed the published results, which were obtained with different corpora, retrievers, and writers. The matched comparison isolates the contribution of reasoning: with the same candidates, reranking, and writing budget, adding reasoning graphs raises citation F1 by 5.3 points on ScholarQA-CS and 5.1 points on ScholarQA-Multi, and improves precision and recall together (\Cref{tab:scholarqa-full}). Reranking alone accounts for a smaller share of the gain, and LKM Graph also exceeds the reranked commercial search API baseline. Knowing how each paper reached its conclusions lets the writer both cite more accurately and support more of what it states.

\subsection{Paper Search and Knowledge-intensive Question Answering}
\label{sec:exp-qa}

We evaluate direct paper retrieval on the 244 queries of PaSaMaster~\citep{du2026towards}, comparing LKM's semantic and hybrid retrieval with a production scientific literature search API by NDCG@$k$. For question answering, GPT-5.4 is fixed as the answering model on ChemBench ($n{=}2{,}785$)~\citep{mirza2025chembench}, PubMedQA ($n{=}500$)~\citep{jin2019pubmedqa}, and SciBench ($n{=}580$)~\citep{wang2024scibench}, and we vary only the retrieval source.

\begin{table}[t]
\centering
\caption{\textbf{Paper search and question answering.} Left: NDCG@$k$ on PaSaMaster. Right: accuracy (\%) on ChemBench, PubMedQA, and SciBench with GPT-5.4 fixed as the answering model.}
\label{tab:search-qa}
\small
\setlength{\tabcolsep}{3.5pt}
\begin{tabular}[t]{@{}lccc@{}}
\toprule
\textbf{Search API} & @5 & @10 & @20 \\
\midrule
Commercial API & 21.44 & 19.77 & 19.28 \\
LKM Semantic & 21.17 & 20.43 & 20.54 \\
LKM Hybrid & \textbf{22.19} & \textbf{21.00} & \textbf{20.93} \\
\bottomrule
\end{tabular}
\hfill
\begin{tabular}[t]{@{}lccc@{}}
\toprule
\textbf{Retrieval} & ChemBench & PubMedQA & SciBench \\
\midrule
None & 64.78 & 77.20 & 76.21 \\
Web search & 70.88 & 81.00 & 86.90 \\
arXiv & 71.14 & 76.60 & 87.07 \\
Commercial API & 72.82 & 81.20 & 88.62 \\
LKM & \textbf{74.08} & \textbf{81.40} & \textbf{90.90} \\
\bottomrule
\end{tabular}
\end{table}

LKM Hybrid achieves the highest NDCG at all three depths (\Cref{tab:search-qa}), showing that reasoning objects also serve as entry points for finding papers. In question answering, LKM retrieval improves the fixed model by 9.3, 4.2, and 14.7 points over no retrieval and gives the highest accuracy on all three benchmarks, although its lead over the commercial API on PubMedQA is a single question and these point estimates carry no paired intervals. The largest gain appears on SciBench, whose problems require applying established scientific relations, consistent with LKM storing such relations as explicit claims and reasoning chains. Together with the preceding results, this traces a consistent picture: recalling reasoning lets agents find more of the relevant evidence, attribute their statements more faithfully, and reason more accurately.

\section{Conclusion and Outlook}
\label{sec:conclusion}

LKM rebuilds the scientific literature as a knowledge foundation for agentic science. It extracts the literature across disciplines into source-grounded reasoning graphs holding about 1.4 billion claims, organizes them into a Scientific Reasoning Landscape of questions, workflows, and evidence, and is updated with every newly published paper. Agents that build on it discover nearly twice as much known evidence for scientific claims, cite their sources more faithfully, and answer knowledge-intensive questions more accurately.

\paragraph{Limitations.} Extraction by language models is imperfect: our audit covers 100 papers and measures hallucination (0.97\%) but not omission, and extraction errors propagate into bindings, clusters, and retrieved context. Evidence carried mainly by figures, tables, or equations may be captured incompletely, and exact-identity binding leaves paraphrased propositions as separate identities linked only by semantic neighborhoods. The landscape views and the Evidence Engine's categorization are not yet directly evaluated; \Cref{sec:exp-evidence} measures evidence retrieval, not categorization. Our comparisons cover retrieval baselines and a matched graph ablation rather than controlled runs of other synthesis or graph-augmented retrieval systems.

\paragraph{Outlook.} Science advances only as fast as its results accumulate. LKM moves the unit of accumulation from the paper to the reasoning object: because questions, workflows, and evidence share common objects, a technique proven on one problem can surface for another, even in fields that name it differently, extending literature-based discovery~\citep{swanson1986undiscovered} from findings to reasoning. The decisive step is to close the loop. When agents return their observations, failures, and conclusions with provenance and verification, negative results that rarely reach print are kept, open questions sharpen, and each investigation starts from what all others have learned. The literature then becomes not an archive that agents read, but a shared, self-correcting body of knowledge that they write together.

\subsection*{Reproducibility statement}

We support reproducibility by documenting the system construction and each
evaluation stage at the level needed to recreate the reported comparisons.
Section~\ref{sec:evaluation} specifies the benchmark datasets and sample sizes,
retrieval depths and budgets, baselines, metrics, fixed answering model, and the
model and decision thresholds used to construct the pooled SciFact-Open
reference set. Section~\ref{sec:reasoning-graphs} describes graph construction,
canonical identifiers, embedding generation, and retrieval, while
Appendix~\ref{app:interfaces} reports the clustering parameters and the public
interfaces through which LKM outputs can be inspected. We use cited public
benchmarks and preserve their original human annotations. Because LKM is a
continuously updated production system, exact retrieval results depend on the
evaluated index snapshot and query time; fixed offline snapshots are identified
where applicable, and live-service results should be interpreted with this
source of variation in mind.

\subsection*{AI use statement}

In this work, we used generative AI tools for writing assistance and language
polishing, retrieval and discovery of related work, assistance with research
ideation and experiment execution, code and figure preparation, and synthetic
dataset generation. Large language models are also components of the system
and its evaluation as described in the paper, including reasoning-graph
extraction (\Cref{sec:reasoning-graphs}) and the answering models, writers,
and judges of the benchmarks (\Cref{sec:evaluation}). The authors reviewed all
AI-assisted text, code, figures, and results, and take responsibility for the
final content of this paper, including its text, claims, and research artifacts.

\bibliographystyle{iclr2027_conference}
\bibliography{references}

@article{asai2024openscholar,journal={arXiv preprint arXiv:2411.14199},title={OpenScholar: Synthesizing Scientific Literature with Retrieval-augmented LMs},author={Asai, Akari and He, Jacqueline and Shao, Rulin and Shi, Weijia and Singh, Amanpreet and Chang, Joseph Chee and Lo, Kyle and Soldaini, Luca and Feldman, Sergey and {D'Arcy}, Mike and Wadden, David and Latzke, Matt and Tian, Minyang and Ji, Pan and Liu, Shengyan and Tong, Hao and Wu, Bohao and Xiong, Yanyu and Zettlemoyer, Luke and Neubig, Graham and Weld, Dan and Downey, Doug and Yih, Wen-tau and Koh, Pang Wei and Hajishirzi, Hannaneh},year={2024}}

@article{ding2025scirag,journal={arXiv preprint arXiv:2511.14362},title={SciRAG: Adaptive, Citation-Aware, and Outline-Guided Retrieval and Synthesis for Scientific Literature},author={Ding, Hang and Zhao, Yilun and Hu, Tiansheng and Patwardhan, Manasi and Cohan, Arman},year={2025}}

@inproceedings{jin2019pubmedqa,booktitle={Proceedings of EMNLP-IJCNLP},title={{PubMedQA}: A Dataset for Biomedical Research Question Answering},author={Jin, Qiao and Dhingra, Bhuwan and Liu, Zhengping and Cohen, William W. and Lu, Xinghua},year={2019},pages={2567--2577},doi={10.18653/v1/D19-1259}}

@inproceedings{karpukhin2020dense,
 title={Dense Passage Retrieval for Open-Domain Question Answering},
 author={Karpukhin, Vladimir and Oguz, Barlas and Min, Sewon and Lewis, Patrick and Wu, Ledell and Edunov, Sergey and Chen, Danqi and Yih, Wen-tau},
 booktitle={Proceedings of EMNLP}, year={2020}, pages={6769--6781}, doi={10.18653/v1/2020.emnlp-main.550}}

@inproceedings{lewis2020retrieval,
 title={Retrieval-Augmented Generation for Knowledge-Intensive NLP Tasks},
 author={Lewis, Patrick and Perez, Ethan and Piktus, Aleksandra and Petroni, Fabio and others},
 booktitle={Advances in Neural Information Processing Systems}, volume={33}, year={2020}, url={https://arxiv.org/abs/2005.11401}}

@article{asai2026synthesizing,
 title={Synthesizing scientific literature with retrieval-augmented language models},
 author={Asai, Akari and He, Jacqueline and Shao, Rulin and Shi, Weijia and others},
 journal={Nature}, volume={650}, pages={857--863}, year={2026}, doi={10.1038/s41586-025-10072-4}}

@article{skarlinski2024language,
 title={Language agents achieve superhuman synthesis of scientific knowledge},
 author={Skarlinski, Michael D. and Cox, Sam and Laurent, Jon M. and Braza, James D. and Hinks, Michaela and Hammerling, Michael J. and Ponnapati, Manvitha and Rodriques, Samuel G. and White, Andrew D.},
 journal={arXiv preprint arXiv:2409.13740}, year={2024}, url={https://arxiv.org/abs/2409.13740}}

@inproceedings{he2025pasa,
 title={{PaSa}: An {LLM} Agent for Comprehensive Academic Paper Search},
 author={He, Yichen and Huang, Guanhua and Feng, Peiyuan and Lin, Yuan and Zhang, Yuchen and Li, Hang and E, Weinan},
 booktitle={Proceedings of ACL}, year={2025}, pages={11663--11679}, doi={10.18653/v1/2025.acl-long.572}}

@article{du2026towards,
 title={Towards Self-Evolving Agentic Literature Retrieval},
 author={Du, Yuwen and Jin, Tian and Kang, Jing and Pang, Xianghe and Chai, Jingyi and Miao, Tingjia and Liu, Fenyi and Wang, WenHao and Yao, Sikai and Zhang, Yuzhi and others},
 journal={arXiv preprint arXiv:2605.14306}, year={2026}, url={https://arxiv.org/abs/2605.14306}}

@inproceedings{lo2020s2orc,
 title={{S2ORC}: The Semantic Scholar Open Research Corpus},
 author={Lo, Kyle and Wang, Lucy Lu and Neumann, Mark and Kinney, Rodney and Weld, Daniel},
 booktitle={Proceedings of ACL}, year={2020}, pages={4969--4983}, doi={10.18653/v1/2020.acl-main.447}}

@article{jaradeh2019orkg,
 title={Open Research Knowledge Graph: Next Generation Infrastructure for Semantic Scholarly Knowledge},
 author={Jaradeh, Mohamad Yaser and Oelen, Allard and Farfar, Kheir Eddine and Prinz, Manuel and D'Souza, Jennifer and Kismihok, Gabor and Stocker, Markus and Auer, Soren},
 journal={arXiv preprint arXiv:1901.10816}, year={2019}, url={https://arxiv.org/abs/1901.10816}}

@article{groth2010nanopublication,
 title={The anatomy of a nanopublication},
 author={Groth, Paul and Gibson, Andrew and Velterop, Jan},
 journal={Information Services \& Use}, volume={30}, pages={51--56}, year={2010}, doi={10.3233/ISU-2010-0613}}

@article{clark2014micropublications,
 title={Micropublications: a semantic model for claims, evidence, arguments and annotations in biomedical communications},
 author={Clark, Tim and Ciccarese, Paolo N. and Goble, Carole A.},
 journal={Journal of Biomedical Semantics}, volume={5}, pages={28}, year={2014}, doi={10.1186/2041-1480-5-28}}

@inproceedings{luan2018multi,
 title={Multi-Task Identification of Entities, Relations, and Coreference for Scientific Knowledge Graph Construction},
 author={Luan, Yi and He, Luheng and Ostendorf, Mari and Hajishirzi, Hannaneh},
 booktitle={Proceedings of EMNLP}, year={2018}, pages={3219--3232}, doi={10.18653/v1/D18-1360}}

@inproceedings{jain2020scirex,
 title={{SciREX}: A Challenge Dataset for Document-Level Information Extraction},
 author={Jain, Sarthak and van Zuylen, Madeleine and Hajishirzi, Hannaneh and Beltagy, Iz},
 booktitle={Proceedings of ACL}, year={2020}, pages={7506--7516}, doi={10.18653/v1/2020.acl-main.670}}

@inproceedings{wadden2020fact,
 title={Fact or Fiction: Verifying Scientific Claims},
 author={Wadden, David and Lin, Shanchuan and Lo, Kyle and Wang, Lucy Lu and van Zuylen, Madeleine and Cohan, Arman and Hajishirzi, Hannaneh},
 booktitle={Proceedings of EMNLP}, year={2020}, pages={7534--7550}, doi={10.18653/v1/2020.emnlp-main.609}}

@inproceedings{kulkarni2018annotated,
 title={An Annotated Corpus for Machine Reading of Instructions in Wet Lab Protocols},
 author={Kulkarni, Chaitanya and Xu, Wei and Ritter, Alan and Machiraju, Raghu},
 booktitle={Proceedings of NAACL-HLT}, year={2018}, pages={97--106}, doi={10.18653/v1/N18-2016}}

@article{teufel2002summarizing,
 title={Summarizing Scientific Articles: Experiments with Relevance and Rhetorical Status},
 author={Teufel, Simone and Moens, Marc},
 journal={Computational Linguistics}, volume={28}, number={4}, pages={409--445}, year={2002}, doi={10.1162/089120102762671936}}

@article{belhajjame2015suite,
 title={Using a suite of ontologies for preserving workflow-centric research objects},
 author={Belhajjame, Khalid and Zhao, Jun and Garijo, Daniel and Gamble, Matthew and others},
 journal={Journal of Web Semantics}, volume={32}, pages={16--42}, year={2015}, doi={10.1016/j.websem.2015.01.003}}

@article{swanson1986undiscovered,
 title={Undiscovered Public Knowledge}, author={Swanson, Don R.},
 journal={The Library Quarterly}, volume={56}, number={2}, pages={103--118}, year={1986}, doi={10.1086/601720}}

@article{edge2024local,
 title={From Local to Global: A Graph {RAG} Approach to Query-Focused Summarization},
 author={Edge, Darren and Trinh, Ha and Cheng, Newman and Bradley, Joshua and Chao, Alex and Mody, Apurva and Truitt, Steven and Metropolitansky, Dasha and Ness, Robert Osazuwa and Larson, Jonathan},
 journal={arXiv preprint arXiv:2404.16130}, year={2024}, url={https://arxiv.org/abs/2404.16130}}

@inproceedings{guo2025lightrag,
 title={{LightRAG}: Simple and Fast Retrieval-Augmented Generation},
 author={Guo, Zirui and Xia, Lianghao and Yu, Yanhua and Ao, Tu and Huang, Chao},
 booktitle={Findings of EMNLP}, year={2025}, pages={10746--10761}, doi={10.18653/v1/2025.findings-emnlp.568}}

@article{mirza2025chembench,
 title={A framework for evaluating the chemical knowledge and reasoning abilities of large language models against the expertise of chemists},
 author={Mirza, Adrian and Alampara, Nawaf and Kunchapu, Sreekanth and Rios-Garcia, Martino and others},
 journal={Nature Chemistry}, volume={17}, pages={1027--1034}, year={2025}, doi={10.1038/s41557-025-01815-x}}

@inproceedings{wang2024scibench,
 title={{SciBench}: Evaluating College-Level Scientific Problem-Solving Abilities of Large Language Models},
 author={Wang, Xiaoxuan and Hu, Ziniu and Lu, Pan and Zhu, Yanqiao and Zhang, Jieyu and Subramaniam, Satyen and Loomba, Arjun R. and Zhang, Shichang and Sun, Yizhou and Wang, Wei},
 booktitle={Proceedings of ICML}, series={Proceedings of Machine Learning Research}, volume={235}, pages={50622--50649}, year={2024}, url={https://proceedings.mlr.press/v235/wang24z.html}}

@article{song2022linking,
  author = {Song, Ningyuan and Cheng, Hanghang and Zhou, Huimin and Wang, Xiaoguang},
  title = {Linking Scholarly Contents: The Design and Construction of an Argumentation Graph},
  journal = {Knowledge Organization},
  volume = {49},
  number = {4},
  pages = {213--235},
  year = {2022},
  doi = {10.5771/0943-7444-2022-4-213}
}

@article{baulin2025discovery,
  author = {Baulin, Vladimir and Cook, Austin and Friedman, Daniel and Lumiruusu, Janna and Pashea, Andrew and Rahman, Shagor and Waldeck, Benedikt},
  title = {The Discovery Engine: A Framework for {AI}-Driven Synthesis and Navigation of Scientific Knowledge Landscapes},
  journal = {arXiv preprint arXiv:2505.17500},
  year = {2025},
  doi = {10.48550/arXiv.2505.17500}
}

@article{vaswani2017attention,
  title = {Attention Is All You Need},
  author = {Vaswani, Ashish and Shazeer, Noam and Parmar, Niki and Uszkoreit, Jakob and Jones, Llion and Gomez, Aidan N. and Kaiser, Lukasz and Polosukhin, Illia},
  journal = {arXiv preprint arXiv:1706.03762},
  year = {2017},
  doi = {10.48550/arXiv.1706.03762}
}

@article{dao2022flashattention,
 title={FlashAttention: Fast and Memory-Efficient Exact Attention with IO-Awareness},
 author={Dao, Tri and Fu, Daniel Y. and Ermon, Stefano and Rudra, Atri and R{\'e}, Christopher},
 journal={arXiv preprint arXiv:2205.14135}, year={2022}, doi={10.48550/arXiv.2205.14135}
}

@article{dao2023flashattention2,
 title={FlashAttention-2: Faster Attention with Better Parallelism and Work Partitioning},
 author={Dao, Tri}, journal={arXiv preprint arXiv:2307.08691}, year={2023}, doi={10.48550/arXiv.2307.08691}
}

@article{wang2020linformer,
 title={Linformer: Self-Attention with Linear Complexity},
 author={Wang, Sinong and Li, Belinda Z. and Khabsa, Madian and Fang, Han and Ma, Hao},
 journal={arXiv preprint arXiv:2006.04768}, year={2020}, doi={10.48550/arXiv.2006.04768}
}

@inproceedings{wadden2022scifactopen,
  title={SciFact-Open: Towards Open-Domain Scientific Claim Verification},
  author={Wadden, David and Lo, Kyle and Kuehl, Bailey and Cohan, Arman and Beltagy, Iz and Wang, Lucy Lu and Hajishirzi, Hannaneh},
  booktitle={Findings of the Association for Computational Linguistics: EMNLP 2022},
  pages={4719--4734},
  year={2022},
  publisher={Association for Computational Linguistics},
  doi={10.18653/v1/2022.findings-emnlp.347}
}

@inproceedings{wadden2022multivers,
  title={MultiVerS: Improving Scientific Claim Verification with Weak Supervision and Full-Document Context},
  author={Wadden, David and Lo, Kyle and Wang, Lucy Lu and Cohan, Arman and Beltagy, Iz and Hajishirzi, Hannaneh},
  booktitle={Findings of the Association for Computational Linguistics: NAACL 2022},
  pages={61--76},
  year={2022},
  publisher={Association for Computational Linguistics},
  doi={10.18653/v1/2022.findings-naacl.6}
}

@article{zhang2025qwen3embedding,
  title={Qwen3 Embedding: Advancing Text Embedding and Reranking Through Foundation Models},
  author={Zhang, Yanzhao and Li, Mingxin and Long, Dingkun and Zhang, Xin and Lin, Huan and Yang, Baosong and Xie, Pengjun and Yang, An and Liu, Dayiheng and Lin, Junyang and Huang, Fei and Zhou, Jingren},
  journal={arXiv preprint arXiv:2506.05176},
  year={2025},
  doi={10.48550/arXiv.2506.05176}
}
\appendix
\section{LKM Interfaces and Agent Integration}
\label{app:interfaces}

\subsection{LKM Website}
The public LKM site (\href{https://lkm.bohrium.com/web/en}{\textbf{Knowledge Explorer}}) searches claims and questions, opens reasoning chains, and shows the
source-paper graph. \href{https://lkm.bohrium.com/web/en/evidence-engine}{\textbf{Evidence Engine}} retrieves evidence that supports
or challenges a research question and produces a cited synthesis.
\href{https://lkm.bohrium.com/web/en/landscape}{\textbf{Science Landscape}} organizes common-question clusters by discipline.
\href{https://lkm.bohrium.com/web/en/workflows}{\textbf{Workflows}} presents reproducible workflow families derived from
reasoning chains. All four operate on source-grounded objects and preserve the
link to the originating paper.

\subsection{Integration Modes}
Agents can access LKM in three ways. The \href{https://s.apifox.cn/33d12311-ec59-4a5c-a849-391704fe7f84}{\textbf{OpenAPI}} provides direct HTTP access and is appropriate for production services and custom orchestration. The \href{https://docs.bohrium.com/docs/bohrctl/install/}{\textbf{bohr CLI}} offers shell-level access for research scripts, batch jobs, and reproducible command-line workflows. \textbf{Bohrium skills} expose the same operations as tool calls inside an agent runtime, handling request construction and response decoding. These modes share the same identifiers and semantics; the choice is an integration concern rather than a different knowledge source.

\subsection{Question-family Construction}
\label{app:question-family-construction}

The clustering input contains 34,413,217 question nodes from arXiv papers,
represented in the shared 512-dimensional embedding space. We construct an
exact mutual $k$-nearest-neighbor graph over all nodes with $k=40$, retaining
edges whose cosine similarity is at least $0.78$; connected components with at
least three nodes form dense cores. Cores whose centroid cosine similarity is at
least $0.84$ are merged into candidate family centers. Each question node is
assigned to at most one center by highest cosine similarity and is accepted when
its similarity reaches the center's adaptive threshold, capped at $0.65$;
otherwise it remains unassigned. Families larger than 2,000 nodes are
recursively reclustered, raising the core similarity threshold by $0.03$ per
level for at most six levels. A final family contains at least four question
nodes from at least three papers. An LLM summarizes each family from its member
questions and papers, producing the shared research question and its principal
variants; the summary remains linked to the member objects.

Restricting to families with at least five nodes yields 988,894 families with
31,910,266 member nodes (92.73\% node coverage), drawn from 2,882,306 of
2,896,083 input papers (99.52\% paper coverage). Family size has mean 32.27,
median 18, 90th percentile 62, 99th percentile 243, and maximum 3,273 nodes; the
number of distinct papers per family has mean 20.20, median 10, 90th percentile
42, 99th percentile 166, and maximum 3,245.

\subsection{LKM OpenAPI}
The public API rooted at \url{https://open.bohrium.com/openapi/v2/lkm} exposes the
documented endpoints summarized below.

\begin{table}[htbp]
\centering
\caption{LKM OpenAPI endpoints and key parameters. \texttt{/papers/parse*} denotes the family of asynchronous endpoints under this prefix that submit a PDF, poll the parse task, and return the extracted graph.}
\label{tab:lkm-openapi}
\footnotesize
\renewcommand{\arraystretch}{1.18}
\begin{tabularx}{\linewidth}{@{}>{\raggedright\arraybackslash}p{0.16\linewidth}>{\raggedright\arraybackslash}p{0.21\linewidth}>{\raggedright\arraybackslash}X>{\raggedright\arraybackslash}p{0.27\linewidth}@{}}
\toprule
\textbf{Interface} & \textbf{Method/path} & \textbf{Purpose} & \textbf{Key parameters}\\
\midrule
Knowledge search & \texttt{POST} \path{/search} & Retrieve abstracts, claims, questions, or chains & \texttt{query}, \texttt{keywords}, \texttt{scopes}, \texttt{retrieval\_mode}, \texttt{filters}, \texttt{offset/limit}\\
Reasoning search & \texttt{POST} \path{/reasoning/search} & Retrieve similar complete reasoning chains & Natural-language query; retrieval mode; paper/date filters; pagination\\
Paper graph & \texttt{POST} \path{/papers/graph} & Return the complete graph extracted from a paper & \texttt{paper\_id}, \texttt{package\_id}, DOI, or title\\
Claim reasoning & \texttt{GET} \path{/claims/{id}/reasoning} & Trace why a known conclusion claim holds & Global conclusion ID (\texttt{gcn\_...})\\
Batch hydration & \texttt{POST} \path{/variables/batch} & Retrieve details for multiple nodes & List of global node IDs\\
Feedback & \texttt{POST} \path{/feedback} & Report a data or service issue & Feedback type, message, optional node/paper ID\\
PDF extraction & \texttt{POST/GET} \path{/papers/parse*} & Submit, poll, and retrieve PDF extraction & PDF payload; parse task ID\\
Citations & \texttt{POST} \path{/papers/citations} & Retrieve references and citing papers & Paper identifier and query options\\
\bottomrule
\end{tabularx}
\end{table}

For \texttt{/search}, valid scopes are \texttt{abstract}, \texttt{claim},
\texttt{premise}, \texttt{conclusion}, \texttt{question}, \texttt{problem},
\texttt{open\_question}, \texttt{subproblem}, and
\texttt{reasoning\_chain}. \texttt{hybrid} combines semantic and lexical
retrieval; \texttt{semantic} favors speed; \texttt{lexical} targets exact
terms. A successful response has \texttt{code=0}; hits are in
\texttt{data.variables}, paper metadata in \texttt{data.papers}, and pagination
is indicated by \texttt{has\_more}. Scores are ranking signals, not scientific
confidence.

The complete schemas, generated examples, authentication details, and error
codes are maintained in the public API reference:\footnote{\url{https://s.apifox.cn/33d12311-ec59-4a5c-a849-391704fe7f84}}.

\section{Additional Results}
\label{app:additional-results}

\Cref{tab:scholarqa-full} reports citation precision, recall, and F1 on ScholarQABench for all configurations behind \Cref{fig:scholarqa}, together with retrieval baselines evaluated in the same pipeline and published systems for which ScholarQABench citation F1 is available. Without retrieval, answers carry no citations and all three scores are zero.

\begin{table}[htbp]
\centering
\caption{\textbf{Citation quality on ScholarQABench} (AttrScore; precision, recall, and F1 in \%). The first two groups use the same OpenScholar-style pipeline with GPT-4o as the writer; published systems are listed with their reported F1. OpenScholar values are taken from the preprint that released ScholarQABench~\citep{asai2024openscholar}; the system is otherwise cited in its journal version~\citep{asai2026synthesizing}.}
\label{tab:scholarqa-full}
\small
\setlength{\tabcolsep}{5pt}
\begin{tabular}{@{}lcccccc@{}}
\toprule
& \multicolumn{3}{c}{\textbf{ScholarQA-CS}} & \multicolumn{3}{c}{\textbf{ScholarQA-Multi}} \\
\cmidrule(lr){2-4}\cmidrule(lr){5-7}
\textbf{System} & P & R & F1 & P & R & F1 \\
\midrule
arXiv search & 33.6 & 35.4 & 34.5 & 35.5 & 35.7 & 35.6 \\
Web search & 24.5 & 29.0 & 26.6 & 29.9 & 35.6 & 32.5 \\
Commercial API & 49.6 & 55.4 & 52.3 & 54.3 & 59.5 & 56.8 \\
Commercial API + rerank & 51.2 & 56.0 & 53.5 & 55.7 & 60.2 & 57.9 \\
\midrule
LKM Search & 48.5 & 50.7 & 49.6 & 51.8 & 53.8 & 52.8 \\
LKM Search + rerank & 52.2 & 54.6 & 53.4 & 55.0 & 57.7 & 56.3 \\
LKM Graph + rerank & \textbf{57.2} & \textbf{60.3} & \textbf{58.7} & \textbf{59.8} & \textbf{63.1} & \textbf{61.4} \\
\midrule
OpenScholar-8B~\citep{asai2024openscholar} & -- & -- & 47.9 & -- & -- & 42.8 \\
OpenScholar-70B~\citep{asai2024openscholar} & -- & -- & 45.9 & -- & -- & 54.7 \\
OpenScholar-GPT-4o~\citep{asai2024openscholar} & -- & -- & 39.5 & -- & -- & 37.5 \\
PaperQA2~\citep{skarlinski2024language} & -- & -- & 48.0 & -- & -- & 47.2 \\
SciRAG-GPT-4o~\citep{ding2025scirag} & -- & -- & 34.0 & -- & -- & 37.8 \\
\bottomrule
\end{tabular}
\end{table}

\end{document}